\documentclass[sigconf]{acmart}

\AtBeginDocument{%
  }

\setcopyright{acmcopyright}
\copyrightyear{2018}
\acmYear{2018}
\acmDOI{XXXXXXX.XXXXXXX}

\acmConference[Conference acronym 'XX]{Make sure to enter the correct
  conference title from your rights confirmation emai}{June 03--05,
  2018}{Woodstock, NY}
\acmISBN{978-1-4503-XXXX-X/18/06}

\usepackage{amsmath,graphicx,amsfonts,amsthm}
\usepackage{algorithmic}
\usepackage{algorithm}
\usepackage{array}
\usepackage{textcomp}
\usepackage{stfloats}
\usepackage{url}
\usepackage{verbatim}
\usepackage{graphicx}
\usepackage{subfigure}
\usepackage{utfsym,colortbl}

\definecolor{mygray}{gray}{.92}

\begin{document}

%%
%% The "title" command has an optional parameter,
%% allowing the author to define a "short title" to be used in page headers.
\title{EGGS: Edge Guided Gaussian Splatting for Radiance Fields}

%%
%% The "author" command and its associated commands are used to define
%% the authors and their affiliations.
%% Of note is the shared affiliation of the first two authors, and the
%% "authornote" and "authornotemark" commands
%% used to denote shared contribution to the research.
\author{Yuanhao Gong}
\email{gong@szu.edu.cn}
\orcid{0000-0001-5702-1927}
\affiliation{%
  \institution{Electronics and Information Engineering, Shenzhen University, China}
  \country{}
}

%%
%% By default, the full list of authors will be used in the page
%% headers. Often, this list is too long, and will overlap
%% other information printed in the page headers. This command allows
%% the author to define a more concise list
%% of authors' names for this purpose.
\renewcommand{\shortauthors}{Yuanhao et al.}

%%
%% The abstract is a short summary of the work to be presented in the
%% article.
\begin{abstract}
The Gaussian splatting methods are getting popular. However, their loss function only contains the $\ell_1$ norm and the structural similarity between the rendered and input images, without considering the edges in these images. It is well-known that the edges in an image provide important information. Therefore, in this paper, we propose an Edge Guided Gaussian Splatting (EGGS) method that leverages the edges in the input images. More specifically, we give the edge region a higher weight than the flat region. With such edge guidance, the resulting Gaussian particles focus more on the edges instead of the flat regions.  Moreover, such edge guidance does not crease the computation cost during the training and rendering stage. The experiments confirm that such simple edge-weighted loss function indeed improves about $1\sim2$ dB on several difference data sets. With simply plugging in the edge guidance, the proposed method can improve all Gaussian splatting methods in different scenarios, such as human head modeling, building 3D reconstruction, etc.
\end{abstract}

%%
%% The code below is generated by the tool at http://dl.acm.org/ccs.cfm.
%% Please copy and paste the code instead of the example below.
%%


%%
%% Keywords. The author(s) should pick words that accurately describe
%% the work being presented. Separate the keywords with commas.
%% A "teaser" image appears between the author and affiliation
%% information and the body of the document, and typically spans the
%% page.
%\begin{teaserfigure}
%  \includegraphics[width=\textwidth]{}
%  \caption{Seattle Mariners at Spring Training, 2010.}
%  \Description{Enjoying the baseball game from the third-base
%  seats. Ichiro Suzuki preparing to bat.}
 % \label{fig:teaser}
%\end{teaserfigure}

%%
%% This command processes the author and affiliation and title
%% information and builds the first part of the formatted document.
\maketitle

\section{Introduction}
Getting 3D signals from multi-view images is a key task in the wide-reaching field of computer vision. It involves the intricate job of examining and interpreting the varying viewpoints presented by the array of images, all to build a precise 3D representation of the subject. This core task underpins a host of applications and studies in computer vision, making it a crucial area to grasp and comprehend.

The challenge here is to understand depth and perspective from two-dimensional images, which is not an easy task. It's more than just viewing images - it's about turning flat visuals into a three-dimensional perspective. This involves advanced math techniques that are complex and sophisticated. Strong algorithms are also needed to manage this translation process. They need to interpret and analyze data quickly and accurately. On top of that, we need significant computational power to process the large amount of data and operations. This makes the task complex and challenging, but also quite interesting from a technical perspective.

\begin{figure}
	\subfigure[multi views]{\includegraphics[width=0.25\linewidth]{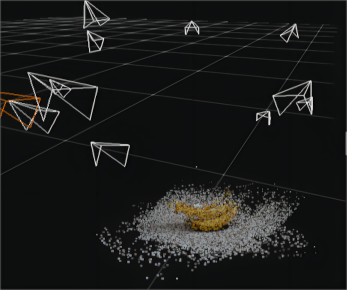}}
	\subfigure[one view]{\includegraphics[width=0.3\linewidth]{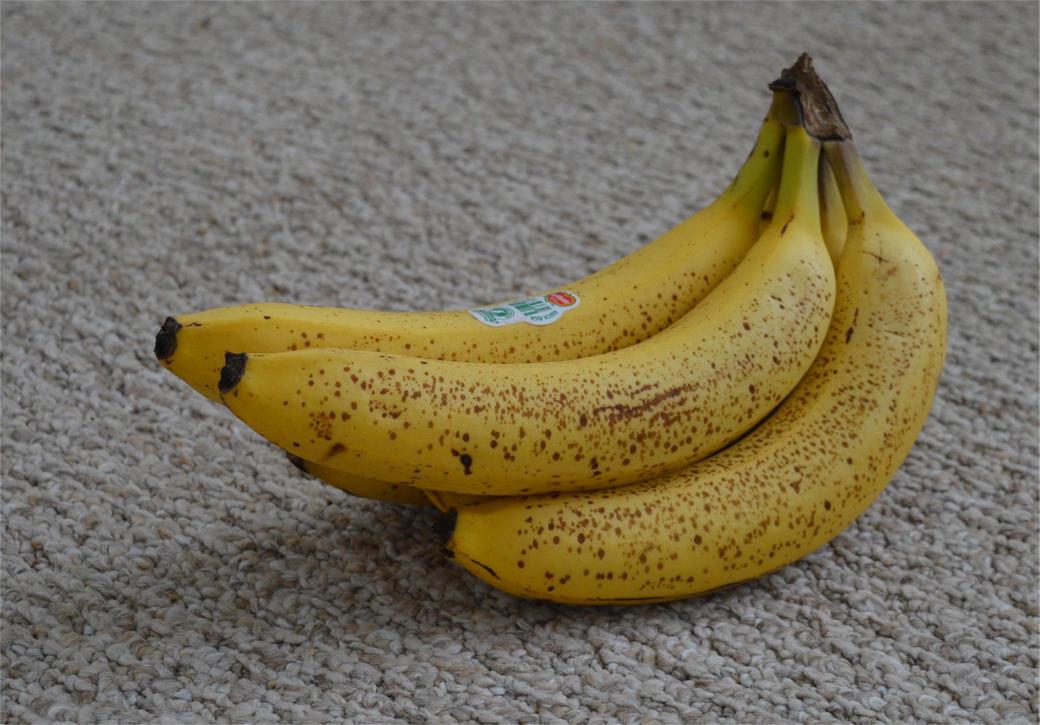}}
	\subfigure[edge guide in this view]{\includegraphics[width=0.3\linewidth]{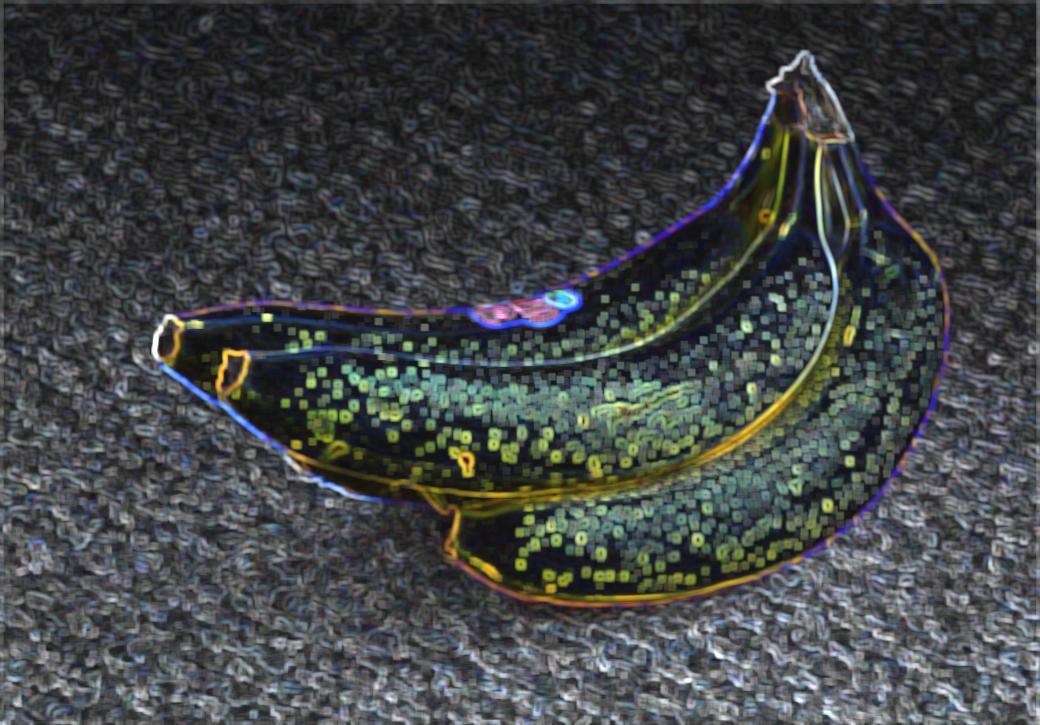}}
	
\hspace{0.25\linewidth}
		\subfigure[another view]{\includegraphics[width=0.3\linewidth]{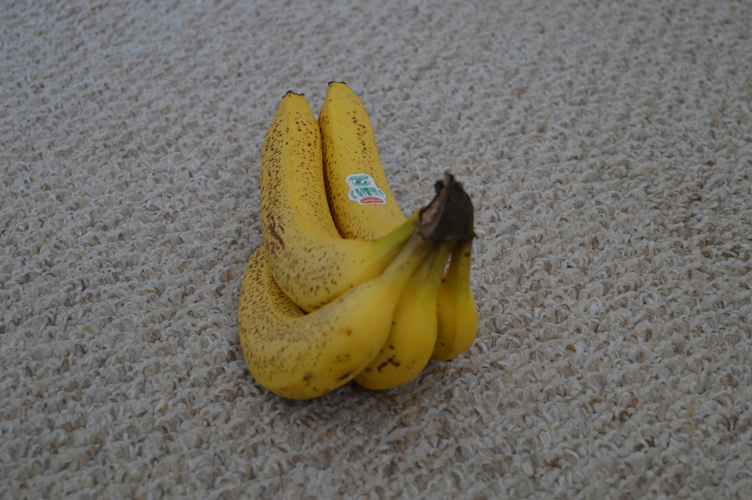}}
	\subfigure[edge guide in this view]{\includegraphics[width=0.3\linewidth]{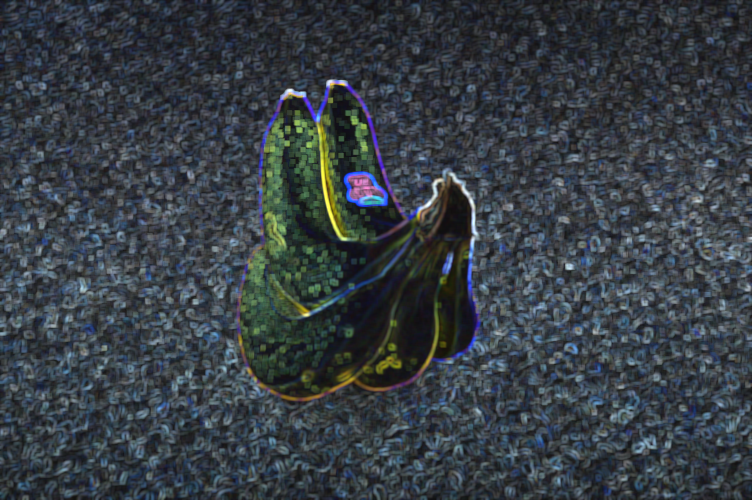}}
	\caption{The edge guide in each view image can force more particles on the edge region, improving the scene accuracy.}
	\label{fig:1}
\end{figure}
NeRF, which stands for Neural Radiance Fields, represents a groundbreaking and innovative method in the field of 3D modelling. It employs the use of a fully connected deep network, a complex and intricate system, to model the volumetric scene function. This function is integral to creating a realistic and immersive 3D environment. What sets NeRF apart is its ability to generate high-quality, novel views of 3D scenes. This is achieved from sparse input views, meaning that even with limited input data, the system can produce detailed and comprehensive 3D scene representations. This illustrates the power and potential of the NeRF method in transforming the way we approach and utilize 3D modelling technology.

Recently, 3D Gaussian splatting has gained quite a bit of traction~\cite{Kerbl2023}. It's a key player in scene estimation and rendering jobs. It uses the Gaussian distribution to figure out the scene's properties, and then uses this information to create detailed and precise illustrations. It avoids the ray tracing in the NeRF methods. Instead, it uses splatting for the image rendering. This method has been a big step in pushing 3D graphics forward.

\subsection{Particle Representation}
For a 3D signal $f(\vec{x})$, where $\vec{x}$ is the spatial coordinate, it can be expressed as a convolution operation with the classical Dirac delta function
\begin{equation}
	f(\vec{x})=\int f(\vec{\tau})\delta(\vec{x}-\vec{\tau})\mathrm{d}\vec{\tau}\,.
\end{equation} Although this is exact, the abstract delta function is not computationally practical.

To improve the computation property, the above equation is relaxed into the follow discrete expression
\begin{equation}
	\hat{f}(\vec{x})\equiv\sum_{k=1}^{K}f(\vec{\tau}_k)W(\vec{x}-\vec{\tau}_k, h_k)V_k\,,
\end{equation}where $\hat{f}$ is the reconstructed signal from this discrete representation, $K$ is the total number of particles, $k$ is the particle index, $W$ is a particle kernel function, $h_k$ is the kernel parameter, and $V_k$ is the volume of the particle.

In most of cases, the multiplication value $f(\vec{\tau}_k)V_k$ can be treated as one variable $A_k$ for convenience reason, leading to the following particle representation
\begin{equation}
	\label{eq:pr}
	\hat{f}(\vec{x})\equiv\sum_{k=1}^{K}A_kW(\vec{x}-\vec{\tau}_k, h_k)\,.
\end{equation} The introduced $A_k$ can carry multiple features, such as mass, temperature, curvature, etc. It is generic for the particle representation.

With such particle representation, we can evaluate the distance between the original signal $f(\vec{x})$ and its reconstruction from the particle representation. More specifically, the distance is
\begin{equation}
	{ L}(f,\hat{f})=\frac{1}{2}\|f(\vec{x})-\hat{f}(\vec{x})\|^2_2\,.
\end{equation}
One important property of particle representation is that this distance can be reduced if more particles are added. This property makes the particle representation flexible and compact.

The gradients of the particle representation with respect to the spatial coordinate, the parameter $h_k$ and $\tau_k$ are
\begin{eqnarray}
	\frac{\partial L}{\partial \vec{x}}&=&(\hat{f}-f)\sum_{k=1}^{K}A_k \frac{\partial W(\vec{x}-\vec{\tau}_k, h_k)}{\partial \vec{x}},\\\
		\frac{\partial L}{\partial h_k}&=&(\hat{f}-f)\sum_{k=1}^{K}A_k \frac{\partial W(\vec{x}-\vec{\tau}_k, h_k)}{\partial h_k}\,,\\
		\frac{\partial L}{\partial \vec{\tau}}&=&(\hat{f}-f)\sum_{k=1}^{K}A_k \frac{\partial W(\vec{x}-\vec{\tau}_k, h_k)}{\partial \vec{\tau}}\,.
\end{eqnarray} These gradients can be used to update the center and shape parameters of the particles.
\subsection{3D Gaussian Splatting}
The 3D Gaussian splatting (3DGS) method and its variants are special cases of the particle representation. More specifically, the 3D Gaussian splatting uses the anisotropic Gaussian kernels in the particle representation
\begin{eqnarray}
	\hat{f}(\vec{x})&=&\sum_{k=1}^{K}A_kG(\vec{\tau}_k,\Sigma_k)\,,\\
	\mathrm{where}\, G(\vec{\tau}_k,\Sigma_k)&=&\exp[-(\vec{x}-\vec{\tau}_k)^T\Sigma_k^{-1}(\vec{x}-\vec{\tau}_k)]\,.
\end{eqnarray}The non negative covariance matrix is $\Sigma=RSS^TR^T$, where $S$ is a diagonal scaling matrix and $R$ is a rotation matrix.

This 3D Gaussian particle is then splatted on the the 2D image plane. The covariance in 2D is computed via
\begin{equation}
	\label{eq:sig2d}
	\Sigma^{2D}=JW\Sigma W^TJ^T\,,
\end{equation} where $W$ is the world-to-camera matrix and $J$ is a local matrix for the projection.

The color $c(u,v)$ at a view is then defined via an alpha blending
\begin{equation}
	c(u,v)=\sum_{k=1}^{K}c_k\alpha_kG^{2D}_k\prod_{j=1}^{k-1}(1-\alpha_jG_j^{2D})\,,
\end{equation}where $c_k$ is a view dependent color, $\alpha_k$ is the transparency, $G^{2D}$ is a 2D Gaussian function with the covariance matrix $\Sigma^{2D}$ in Eq.~\eqref{eq:sig2d}. And $k$ is sorted from the view direction. The coordinate $(u,v)$ indicates the image space coordinate.

The rendered image $c(u,v)$ is then compared with the observed image $im$ via a $\ell_1$ distance and the structural similarity distance
\begin{equation}
	\label{eq:origloss}
	Loss(c,im)=(1-\lambda)|c-im|+\lambda D_{SSIM}(c,im)\,,
\end{equation} where $\lambda>0$ is a weight parameter and $D_{SSIM}$ is a distance measurement using $SSIM$.

\subsection{Variants of Gaussian Splatting}
Thanks to the splatting, the Gaussian splatting methods do not need the ray tracing to render the observed image at a given view. As a result, these methods are much faster than the Nerf based approaches that require the ray tracing to perform the image rendering.

With the advantages of computational efficiency and the resulting high quality rendered images, various Gaussian splatting methods have been developed. For example, it can be applied in street modeling~\cite{yan2023streetgaussians}, human head modeling~\cite{wang2024gaussianhead,zhou2024headstudio} and human body modeling~\cite{abdal2023gaussian}. In~\cite{gong2024isotropic}, isotropic Gaussian function is adopted to reduce the orientation issue. And A compression method is developed to remove the unimportant Gaussian particles, reducing the file size~\cite{lee2023compact,fan2024lightgaussian}. When we are writing this paper, 2D Gaussian splatting method is developed in~\cite{huang20242d}, where 2D disks are attached to a surface. Similar idea is also shown in~\cite{guedon2023sugar}. This method is more suitable for surface representation instead of volume representation. A survey on 3D Gaussian splatting can be found in~\cite{chen2024survey}.

\subsection{Our Motivation and Contributions}
Different from previous work that focuses on a specific field, we notice that the loss function treats each pixel equally. However, it is well-known that the edges in the image are more important than the flat region. This motivates us to develop an edge guided loss function. Our contributions are
\begin{itemize}
	\item  we introduce the edge guidance for the Gaussian splatting method.
	\item the edge guidance is generic and various edge functions can be adopted.
	\item we valid the effectiveness of the edge guidance via several numerical experiments.
\end{itemize}

\section{Edge Guided Gaussian Splatting}
In the loss function Eq.~\eqref{eq:origloss}, the pixels at edges  and the pixels at flat regions have the same weight. Therefore, these two types of pixels have the same influence on the result. However, we know that the edge information is more important in the human vision perception. It is reasonable to increase the weight for pixels at edges. This motivates us to develop an edge guided Gaussian splatting method, which achieves better results.

\subsection{Edge Guidance}
There are many different edge indicators, such as the classical gradient, Laplacian of Gaussian and Canny detector. These methods can be efficiently evaluated, although the accuracy may not achieve the state-of-the-art. Thanks to the efficiency, they can be deployed in real-time applications.

The edges can also be found by deep neural networks, such as~\cite{Xie2015,Liu2019c,Pu2022}. Thanks to the powerful neural networks, these methods can get a higher accuracy in edge detection. Althoug the accuracy is satisfying, these methods require a large mount of computation resource, such as memory and high performance GPUs. This limits their deployment in practical applications. A survey about edge detection can be found in~\cite{Jing2022}.

In this paper, we use a gradient based edge indicator function. More specifically, we define
\begin{equation}
	\label{eq:edge}
	\phi(u,v)=1+\beta\|\nabla im(u,v)\|_p\,,
\end{equation} where $\beta>0$ is a scalar parameter, $\nabla$ is the gradient operator, $p=\{1,2\}$ indicates the standard $\ell_1$ or $\ell_2$ norm. The parameter $\beta$ controls the gradient influence. If $\beta=0$, then $\phi=1$ and this becomes the original Gaussian splatting method. At the flat region, $\|\nabla im\|\approx 0$, $\phi\approx 1$. At the edges, $\|\nabla im\|>> 0$, leading to $\phi>> 1$. Therefore, this weight function will force Gaussian splatting to have a higher accuracy at the edges, which provides more visual information. One example of this weight function is shown in Fig.~\ref{fig:1}, where the values are linearly scaled for better visualization.

\subsection{Our Loss Function}
We impose the above weight function into the loss function, leading to the edge guided loss that is
\begin{equation}
	\label{eq:ourloss}
	Loss(c,im)=(1-\lambda)\|\phi(u,v)(c-im)\|_1+\lambda D_{SSIM}(c,im)\,.
\end{equation} 

The proposed loss function has several advantages. First, the weight function $\phi(u,v)$ is very generic and other edge indicating functions can also be adopted without changing the pipeline of the proposed method. 

Second, the weight function is only determined by the observed input images. It is independent from the 3D radiance field. It works in the image space rather than in the 3D object space. Therefore, it can be pre-computed without estimating the unknown 3D signal. In contrast, some previous methods such as~\cite{cheng2024gaussianpro,huang20242d} also use regularization on the estimated radiance field, for example, the smoothness of the signal or the normal consistency of the surface. Such smoothness relies on the estimated signal or surface, which has to be updated every step during the training process. 

Third, the weight function does not increase the computation cost. Its computational performance is exactly the same as the original 3DGS. In other words, the proposed method does not increase the computation cost, but can improve the accuracy of the radiance field. The proposed method is a Gaussian splatting method with edge guidance. Thus, we name this method as edge guided Gaussian splatting (EGGS).
\subsection{Others}
The optimization and rendering in the proposed EGGS are exactly the same as the original 3DGS. We use the same splitting, merging and deleting strategy as the 3DGS. The only difference is the proposed edge guidance in the loss function.
\section{Experiments}
With the simple edge guidance defined in Eq.~\eqref{eq:edge} and the corresponding loss function in Eq.~\eqref{eq:ourloss}, we can optimize the Gaussian particles to obtain the scene representation. We test the proposed EGGS on three different data sets and compared the results with the counterpart from the 3DGS method.

\begin{figure}
	{\includegraphics[width=0.8\linewidth]{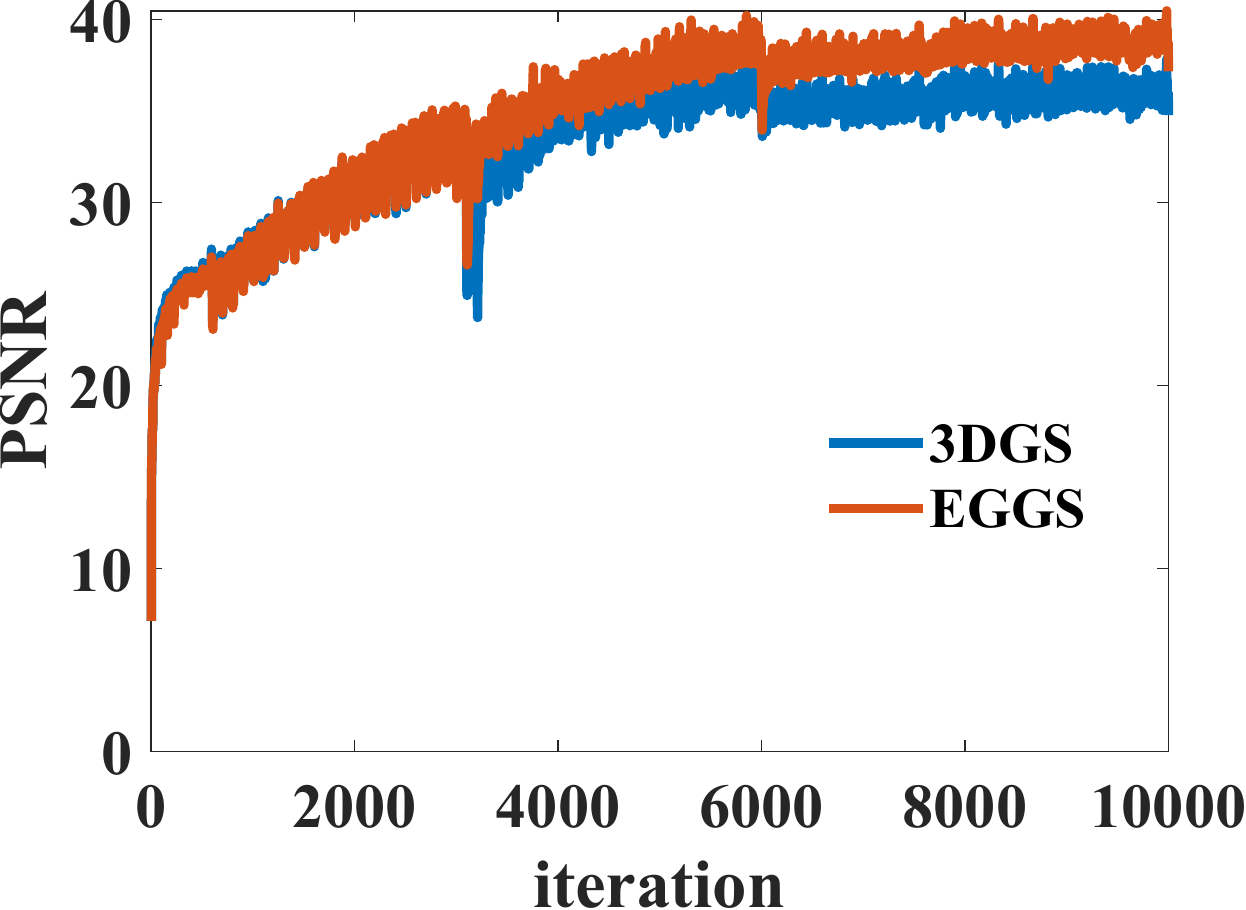}}
	\caption{The PSNR during the training process (the lines are smoothed for better visualization). The blue line is 3DGS and the red line is EGGS. The EGGS can achieve better PSNR.}
	\label{fig:2}
\end{figure}
\begin{figure}
	\subfigure[3DGS]{\includegraphics[width=0.45\linewidth]{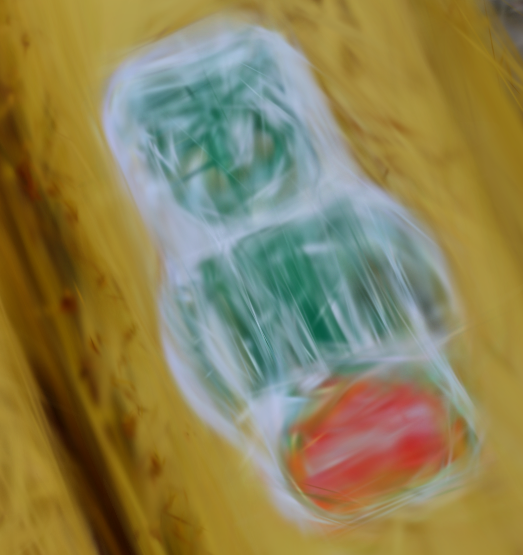}}\hspace{0.05\linewidth}
	\subfigure[EGGS]{\includegraphics[width=0.44\linewidth]{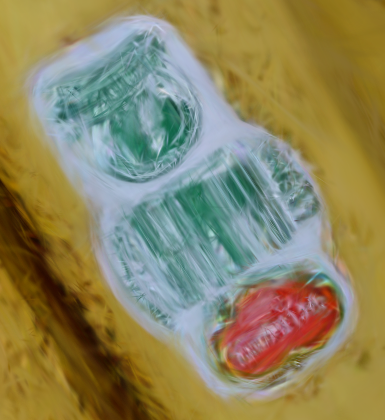}}
	\subfigure[3DGS]{\includegraphics[width=0.45\linewidth]{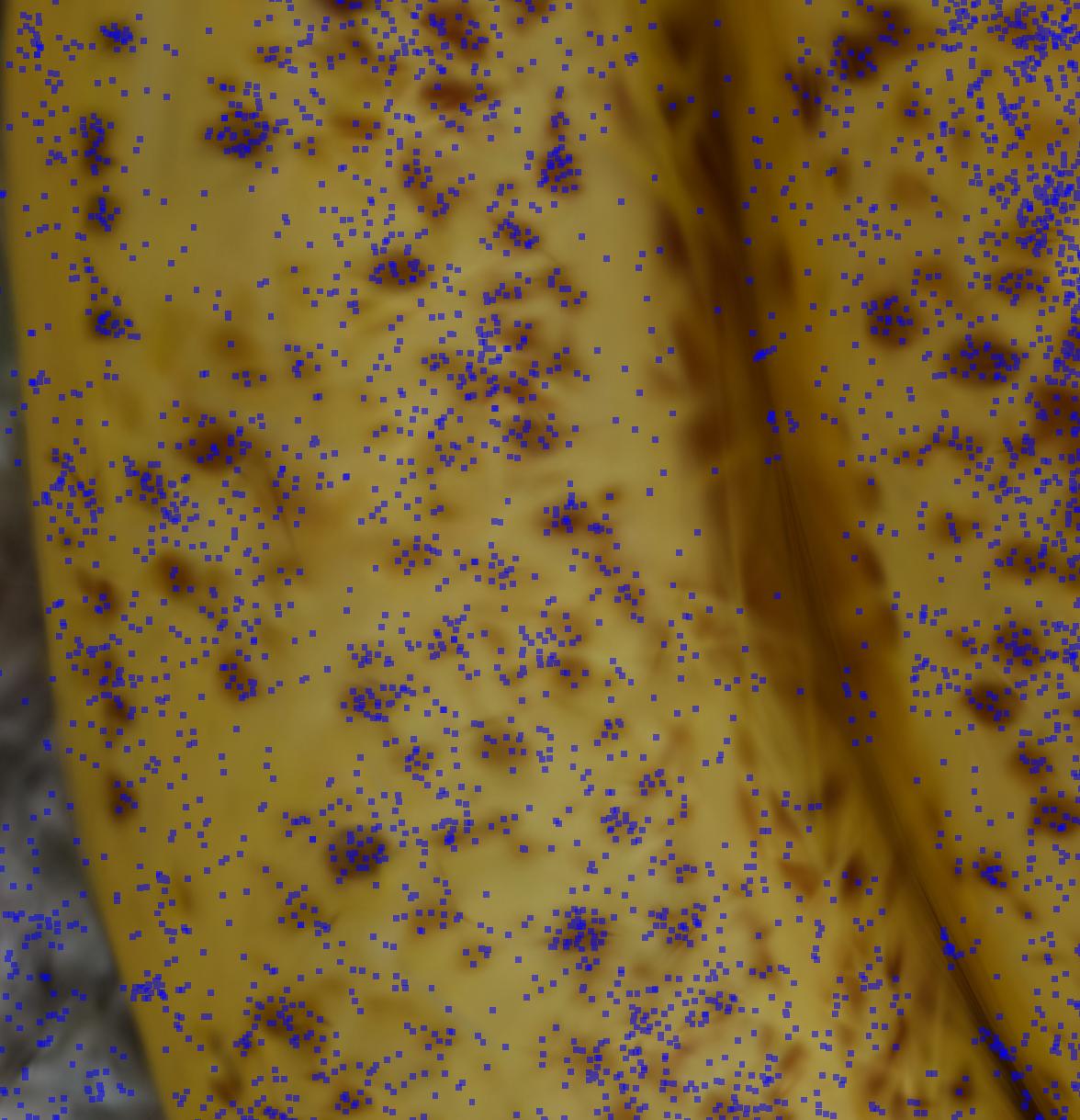}}\hspace{0.02\linewidth}
	\subfigure[EGGS]{\includegraphics[width=0.48\linewidth]{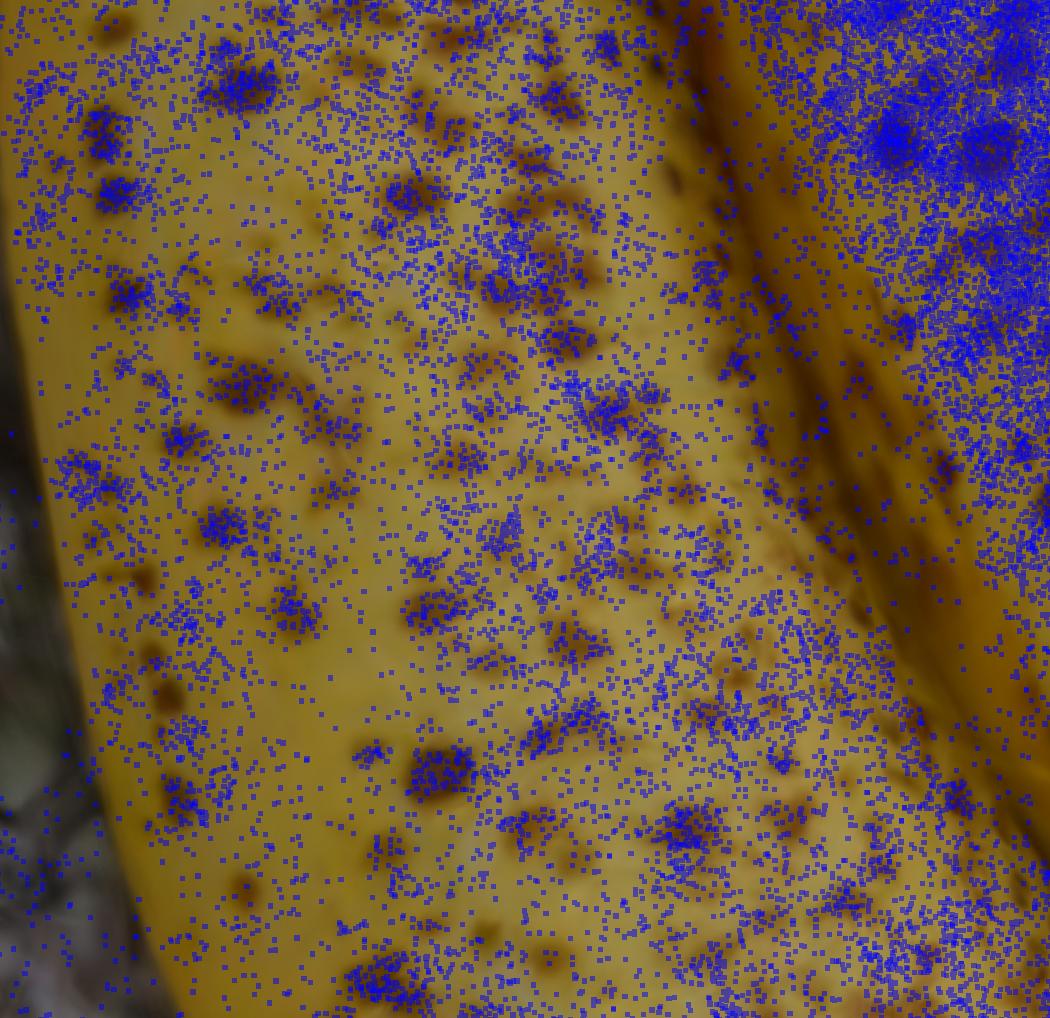}}
	\caption{The detailed difference between 3DGS and EGGS on the banana data set. The top row shows that the edges in the EGGS are much clearer than the edges in 3DGS. The bottom row shows that EGGS puts more particles near the edges.}
	\label{fig:3}
\end{figure}

\subsection{Banana Dataset}
The banana data set contains 16 images at different views. And each image has the $3008\times2000$ resolution. Such high resolution can capture the details in the scene and improve the quality of the radiance field. We use 3DGS and EGGS to perform the reconstruction. The results are shown in Fig.~\ref{fig:2} and \ref{fig:3}.

In Fig.~\ref{fig:2}, we compared the 3DGS (blue line) and EGGS (red line) on this data set. The largest PSNR for 3DGS is about 41.7dB. In contrast, EGGS can achieve 43.8dB, which is about 2.1dB improvement. Such improvement is a big step to improve the radiance field. Be aware that such improvement is only caused by the edge guidance.

We visually show the results in Fig.~\ref{fig:3}, where the left column is for 3DGS and the right is for EGGS. It can be confirmed that EGGS has sharper edges and is much clearer. The letters in the red region are blurred in the 3DGS result and almost invisible. In the EGGS result, these letters can be roughly seen. In the bottom row, we show the RGB images with the Gaussian particle centers (indicated by the blue dots). In 3DGS, the dots are more uniform distributed. In EGGS, the dots are aligned with the edges. Such behavior confirms that the proposed edge guidance indeed forces the Gaussian particle to be aligned with the edges.

\subsection{Train Dataset and Truck Dataset}
\begin{figure}
	\subfigure[train]{\includegraphics[width=0.45\linewidth]{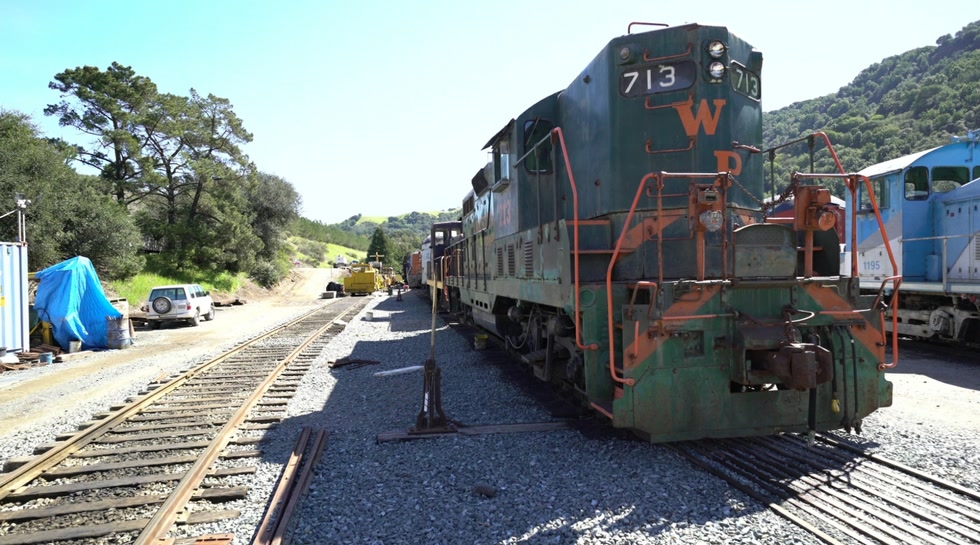}}
	\subfigure[train edge guidance]{\includegraphics[width=0.45\linewidth]{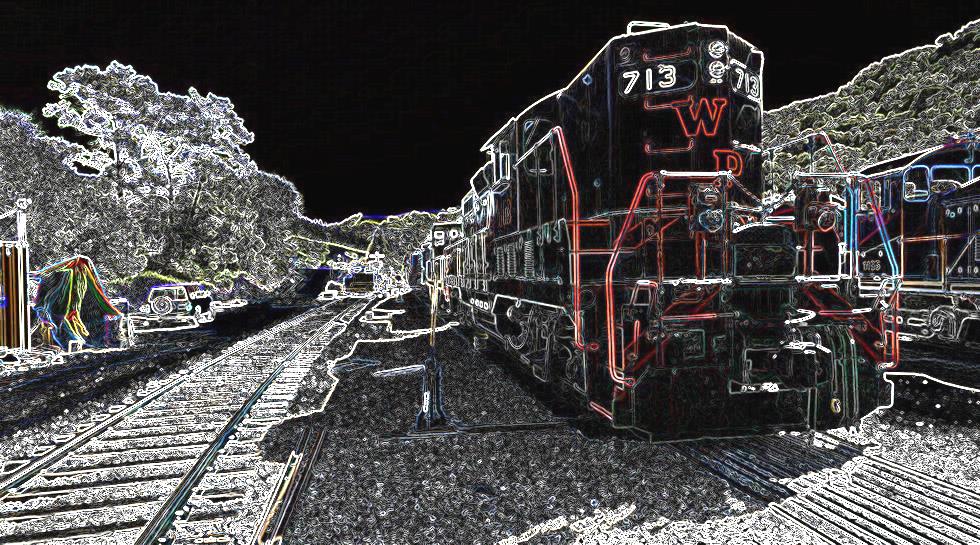}}
	\subfigure[truck]{\includegraphics[width=0.45\linewidth]{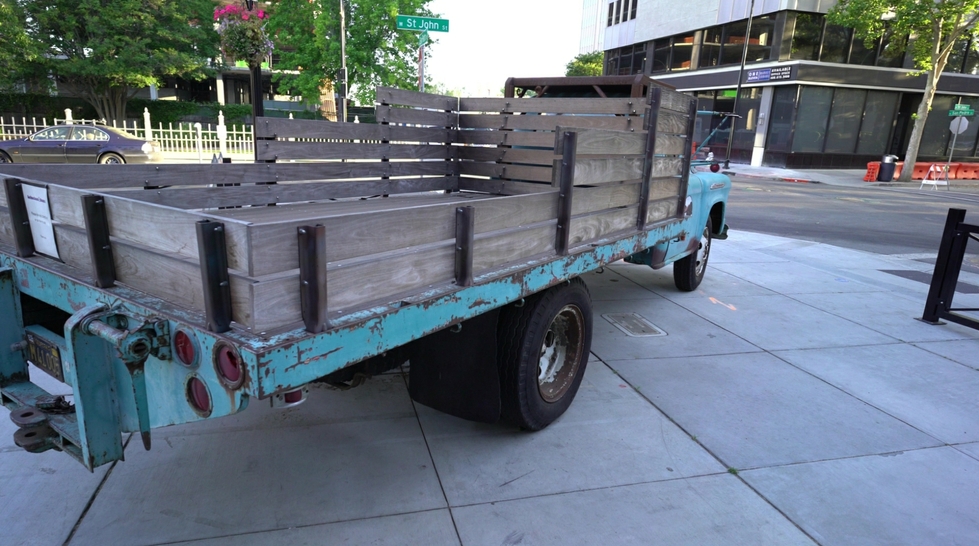}}
	\subfigure[truck edge guidance]{\includegraphics[width=0.45\linewidth]{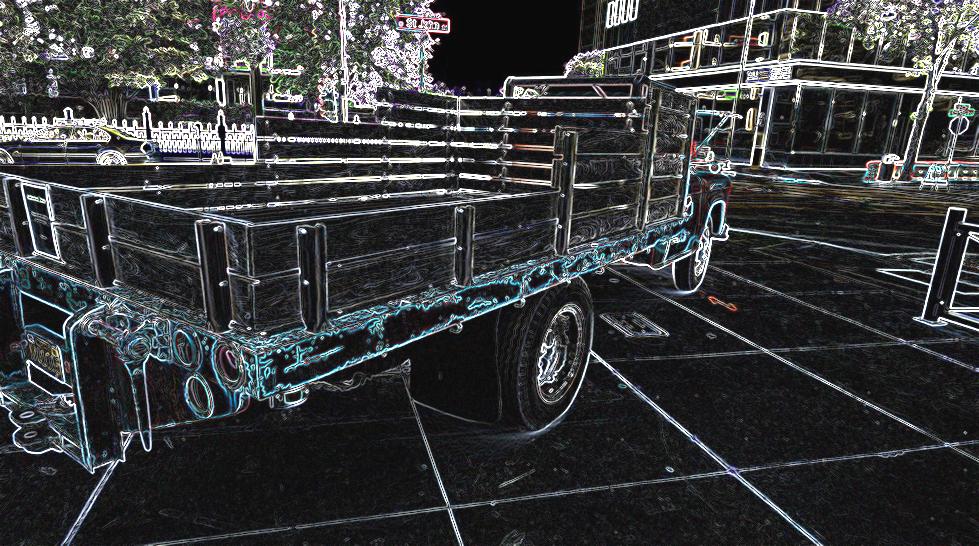}}
	\caption{The edges guidance for the train and truck data set.}
	\label{fig:5}
\end{figure}

\begin{figure}
	{\includegraphics[width=0.8\linewidth]{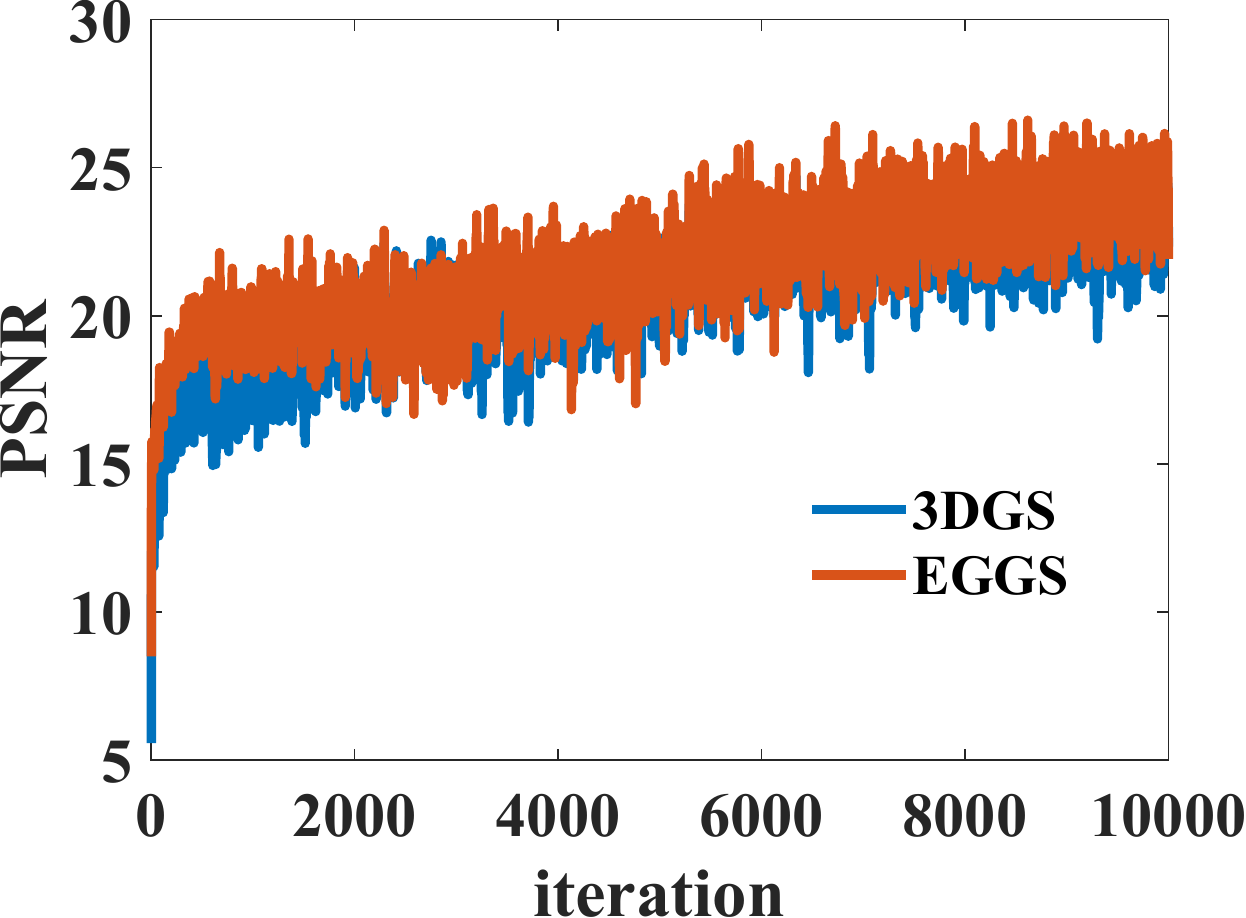}}
	\caption{The PSNR during the training process (the lines are smoothed for better visualization). The blue line is 3DGS and the red line is EGGS. The proposed EGGS can achieve better results on the train data.}
	\label{fig:4}
\end{figure}

The train data set contains 301 images, which have the $980\times545$ resolution. One example input image is shown in Fig.~\ref{fig:5}(a), along with its edge guidance in Fig.~\ref{fig:5}(b). The train process is shown in Fig.~\ref{fig:4} for the 10k iterations. Although it is not the converged state, we can still tell that the edge guidance indeed helps in improving the quality of the estimated radian field. The best PSNR for 3DGS and EGGS on this data set is 28.0 and 29.2, respectively. The improvement is about 1.2dB. 

The truck data set contains 251 images, which have the $979\times546$ resolution. One example input image is shown in Fig.~\ref{fig:5}(c), along with its edge guidance in Fig.~\ref{fig:5}(d). It shows similar behavior as previous data sets. And thus the training curves are omitted here. The best PSNR for 3DGS and EGGS on this data set is 28.4 and 29.5, respectively. The improvement is about 1.1dB. 

These experiment results confirm that the proposed edge guidance indeed can improve the accuracy of the radiance field that is represented by 3DGS. The PSNR gain might depend on the scene.
\subsection{PSNR Improvement}
The PSNR improvements on these data sets are summarized in Table~\ref{table}. In general, the edge guidance can improve the accuracy of the Gaussian particle representation.

The improvement might depend on several things, such as the input image number (view number) , image resolution, the complexity of the 3D scene, the light condition, etc. The improvement in the banana data set is high because the input images have a high resolution and the scene has simple geometry. Most of the edge guidance is on the banana itself. In contrast, the edge guidance in the train and truck data set contains the trees and buildings, which might hamper the edge guidance. 
\begin{table} 
	\centering \caption{The PSNR comparison of EGGS and 3DGS. }
	\begin{tabular}{c|c|c|c}\hline
		\hline
		&Banana&Train&Truck\\
		&$3008\times2000$&$980\times545$&$979\times546$\\
		&16&301&251\\
		\hline
		3DGS &41.7 &28.0 & 28.4\\
		EGGS &\textbf{ 43.8} &\textbf{29.2 } & \textbf{29.5}\\
		\hline
		\rowcolor{gray!20} improved &2.1  &1.2 & 1.1\\
		\hline
	\end{tabular}
	\label{table}
\end{table}

\section{Conclusion}
In this paper, we present a simple yet effective edge guidance for the Gaussian splatting method. The proposed edge guidance is only determined by the multi view input images. And thus it does not increase the computation cost during the training and rendering stage in the Gaussian splatting method. Moreover, as shown in the experiments, this edge guidance can improve the accuracy of the radian field about $1\sim 2$ dB.%, leading to a much clearer rendering results, especially at edges. %Such improvement is a big step for the view synthesis because the edges contain more visual information. 

The proposed edge guidance forces the Gaussian particles to be aligned with the edges in the scene. Therefore, it will help in improving the accuracy of geometry representation~\cite{gong2009symmetry,Gong2023d,Gong2023e}.

The proposed edge guided Gaussian splatting method can achieve higher accuracy for the scene representation and rendering. It can be applied in a large range of applications where edge information is important ~\cite{Gong2012,Yu2019,gong2013a,Yin2019a,gong:phd,Yu2022a,gong:cf,Zong2021,Gong2018,Gong2018a,GONG2019329,Yin2019b,Gong2019a,Gong2019,Gong2019c,Gong2022,Yin2020,Gong2020a,Tang2023a,Gong2021a,Gong2021,gong2024eggs}.
\begin{acks}
	This work was supported by National Natural Science Foundation of China (61907031) and Shenzhen Science and Technology Program (20231121165649002 and JCYJ20220818100005011)
\end{acks}
%%
%% The next two lines define the bibliography style to be used, and
%% the bibliography file.
\bibliographystyle{ACM-Reference-Format}
\bibliography{../../IP}

\end{document}